%% file: iclr2027_conference.tex
\documentclass{article} 
\usepackage{iclr2027_conference,times}

\input{math_commands.tex}

\usepackage{hyperref}
\usepackage{url}
\usepackage{booktabs}
\usepackage{graphicx}
\usepackage{subfigure}
\usepackage{caption}
\usepackage{amssymb}
\usepackage{multirow}

\title{Beyond Mode Collapse: Generating Diverse Synthetic Expert Conversations via Generative Flow Networks}

\author{Sumit Asthana \\
University of Michigan \\
Ann Arbor, MI, USA \\
\texttt{asumit@umich.edu}
\And
Michael Ion \\
University of Michigan \\
Ann Arbor, MI, USA \\
\And
Kevyn Collins Thompson \\
University of Michigan \\
Ann Arbor, MI, USA \\
}

\newtoggle{comments}

\togglefalse{comments}

\newcommand{\sumit}[1]{
\iftoggle{comments}{
\textcolor{violet}{\textbf{(SA:)} #1}}{}
}

\newcommand{\mmdb}{MMDB}
\newcommand{\esconv}{ESConv}

\newcommand{\gemini}{gemini-2.5-flash}
\newcommand{\gflowllm}{GFlow+LLM}
\newcommand{\rlllm}{RL+LLM}
\newcommand{\llmonly}{LLM-only}
\iclrfinalcopy 
\begin{document}

\maketitle

\begin{abstract}
High quality synthetic data is central to post-training LLMs for adaptive-AI applications that represent the diverse expert strategies and decisions in conversations. Prompting LLMs directly or conditioning them on end-use scenarios yields low-diversity data that collapses onto dominant modes.  We propose a method to generate diverse high-quality synthetic data using Generative Flow Networks (GFlowNets). We show that training GFlowNets to generate latent conversation structure using a Gaussian-mixture density over key interaction features (e.g., confusion-episode dynamics, scaffolding–directive balance) enables sampling expert strategies in proportion to their prevalence in the training data. Across two structurally distinct domains, tutoring and emotional-support dialogues, our GFlow-based synthetic data generation approach offers a better balance of fidelity, mode-coverage and authenticity than reinforcement-learning and end-to-end LLM baselines, without copying training data. Evaluated on three downstream outcome prediction tasks, classifiers trained on synthetic GFlowNet-generated conversations provide a stronger training signal than competitive synthesis baselines.

\end{abstract}

\section{Introduction}
Building adaptive AI applications to support tasks involving experts (e.g., answering student questions, providing emotional support) requires understanding the decisions that experts make as conversations unfold. These decisions might include taking actions for probing knowledge, explaining, redirecting, and resolving~\citep{chi2014nature}. As LLMs are increasingly deployed in these domains, including autonomous agentic workflows~\citep{kwan-etal-2024-mt,sun2025training}, reproducing expert decisions faithfully for post-training and evaluation becomes essential. Yet real expert conversations are typically difficult and expensive to obtain; they may also be privacy-sensitive and ultimately cannot be collected at the scale these systems require~\citep{de2023guide,nikolenko2019synthetic}.  

Synthetic conversations offer a scalable alternative for post-training LLMs~\citep{kurakin2023harnessing}. While current LLM-based methods can reproduce plausible, fluent surface language of expert dialogues, they struggle to faithfully reproduce the deeper structure underlying diverse, adaptive decisions that drive a long goal-directed conversation~\citep{laban2026llms}. Prompting with example scenarios or seed data can add surface diversity, such as varied topics~\citep{soudani2026recentconversations}, but the underlying decision patterns can still collapse onto the dominant modes of interaction in the data~\citep{o2024attributing,hamilton-2024-detecting}. Attempted to steer data generation with a hand-crafted proxy quality reward can optimize synthesis toward the dominant highest-scoring mode, suppressing the very diversity that downstream post-training (e.g., RLHF) relies on. 

Recent advances in measuring structural and cognitive properties of synthetic conversations (e.g., expert decisions, challenges)~\citep{ion2026measuring} and factored conversation generation~\citep{macina-etal-2023-mathdial} have opened new opportunities to control the latent structure of a conversation directly, and to measure whether the resulting synthetic conversations exhibit the intended expert decisions and cognitive dynamics~\citep{ion2026measuring}. In particular, Generative Flow Networks (GFlowNets)~\citep{bengio2021flow} provide a controllable way to achieve mode coverage by sampling discrete sequences in proportion to their rewards. In this work, we leverage the capabilities of GFlowNets to train and sample diverse conversation sequences in latent space using a mode-specific Gaussian Mixture Model (GMM). We pose the following research questions.

\begin{itemize}
    \item [RQ1:] Do the capabilities of GFlow networks flexibly address the needs for diverse synthetic data generation?
    \item [RQ2:] How does data generated using GFlowNets compare against baselines on fidelity, coverage and authenticity?
    \item [RQ3:] Does GFlowNet-generated data improve downstream task performance in structured expert conversations?
\end{itemize}

We evaluate our GFlow-based synthetic data generation~\footnote{https://github.com/codez266/diverse-synthetic-expert-conversations} against competitive Reinforcement Learning (RL) and LLM-only generation baselines in two expert assistance domains: (1) math-tutoring and (2) emotional support dialogues. We use multidimensional measures of synthetic data quality that include distributional fidelity, mode coverage, and authenticity with respect to real conversations. We show that GFlow-based synthetic data generations yield higher fidelity expert conversations, while maintaining coverage of diverse expert strategies, without plagiarizing real training conversations. We also evaluate the utility of using synthetic data for downstream outcome prediction tasks. We find that the conversation diversity achieved from using GFlowNet-generated training data provides a stronger training signal than baseline synthetic data approaches.



\section{Background}
\label{sec:background}
Mode collapse in pretrained LLMs is the phenomenon of generating data with limited linguistic and content diversity. Post-training on diverse synthetic data is one way to push a model beyond this distribution~\citep{o2024attributing}, but only if the synthetic data is itself diverse. Extensive work on multi-turn conversations shows that LLMs degrade as conversations lengthen~\citep{kwan-etal-2024-mt}. However, such evaluations have  largely focused on capabilities and turn-level quality: benchmarks probe recollection, refinement, and follow-up and find, while synthetic-text toolkits standardize fluency, utility, privacy, and distributional metrics such as FID and MAUVE~\citep{ramesh-etal-2025-synthtexteval}.

Surface-level interventions such as temperature scaling and scenario prompting can increase variation but tradeoff other quality aspects and limited by LLM's and practitioner's abilities~\citep{nguyen2025turning,chang2024quality}. Real expert conversations are available only in small quantities and cannot be collected at the scale post-training requires. Factoring conversations in a latent move space and then realizing the moves into dialogues provides a useful alternative to control conversation generation~\citep{macina-etal-2023-mathdial, soudani2026recentconversations}. Generative Flow Networks~\citep{bengio2021flow} were introduced to learn a policy that samples discrete, compositional objects with probability proportional to a terminal reward. They provide a promising alternative to reward-maximizing approaches to represent and generate conversations with diverse expert trajectories.

The utility of synthetic data for post-training concentrates in the rare, difficult, and safety-relevant regions, like escalating disclosures, adversarial or multi-intent requests, and extended unresolved confusion that likelihood-driven generation samples least visit~\citep{tan-lee-2025-unmasking}. Recovering this coverage by hand-enumerating persona or scenario taxonomies~\citep{zamfirescu2023prompt} bounds diversity by the designer's anticipation and does not scale to the combinatorial, sequentially dependent space that scenario factors induce, leaving the tail effectively unreachable~\citep{novak2026prompting}. This motivates sampling compositional structure with non-negligible probability on low-density regions, rather than enumeration or unmodified base-model sampling.

\input{3_method}

\section{Evaluations}
We now describe the datasets, features, and metrics for evaluating our proposed GFlow-based conversation synthesis approach. Our empirical results comprise results based on both task-independent, multidimensional dataset quality measures and task-specific training effectiveness.

\subsection{Datasets and Dataset Features}
\label{sec:datasets}
We use two expert datasets: (1) Math Mentor Database (\mmdb)~- a dataset of goal-directed math tutoring conversations, where a tutor is addressing a student's misconception with regards to specific math problems~\citep{ion2026measuring}, and (2) Emotional Support Conversations (\esconv)~- a dataset of emotional support conversations where a support seeker interacts with a support provider regarding a specific emotional issue~\citep{liu-etal-2021-towards}.

We represent each real or synthetic conversation $x$ as a fixed-length vector
$\phi(x)\in\mathbb{R}^d$ of scalar statistics computed from its sequence of (role,
move, utterance) turns. We borrow the feature taxonomy from~\citep{ion2026measuring},
who define features along five dimensions: 1) \emph{surface} features describe
stylistic form independent of content (message-length statistics, turn counts), 2)
\emph{content} features are per-move occurrence rates over the full move/strategy
vocabulary (e.g.\ scaffolding or direct-guidance rates for tutoring, question or
suggestion rates for ESConv), 3) \emph{structure} features aggregate those rates into
coarser category shares (e.g.\ tutor-academic vs.\ socio-emotional turns, or empathic
vs.\ directive strategies), 4) \emph{dynamics} features capture how content changes
over the conversation (e.g.\ confusion-episode duration, the shift in scaffolding
density from the first to the last third, or the position of the first suggestion),
and 5) \emph{outcome} features are binary or scalar end-state indicators (e.g.\
whether the session resolved, whether a breakthrough occurred).
Table~\ref{tab:dataset-summary} summarizes the two source datasets; the full
per-dataset feature list is in Appendix~\ref{app:tab:feature-sets}.

\begin{table}[h]
\centering
\begin{minipage}[t]{0.62\textwidth}
\vspace{0pt}
\centering
\footnotesize
\begin{tabular}{@{}l p{0.3\textwidth} p{0.45\textwidth}@{}}
\toprule
& \textbf{MathMentorDB} & \textbf{ESConv} \\
\midrule
Domain & 1:1 math tutoring & emotional-support dialogue \\
Roles & tutor, student & supporter, seeker \\
Move vocab. & 24 & 9\\
Train convs. & 2{,}500 & 1{,}040 \\
Held-out convs. & 1{,}000 & 260 \\
Avg.\ turns/conv. & 23.8 & 29.5 \\
\bottomrule
\end{tabular}
\end{minipage}
\hfill
\begin{minipage}[t]{0.32\textwidth}
\vspace{0pt}
\centering
\includegraphics[width=\linewidth]{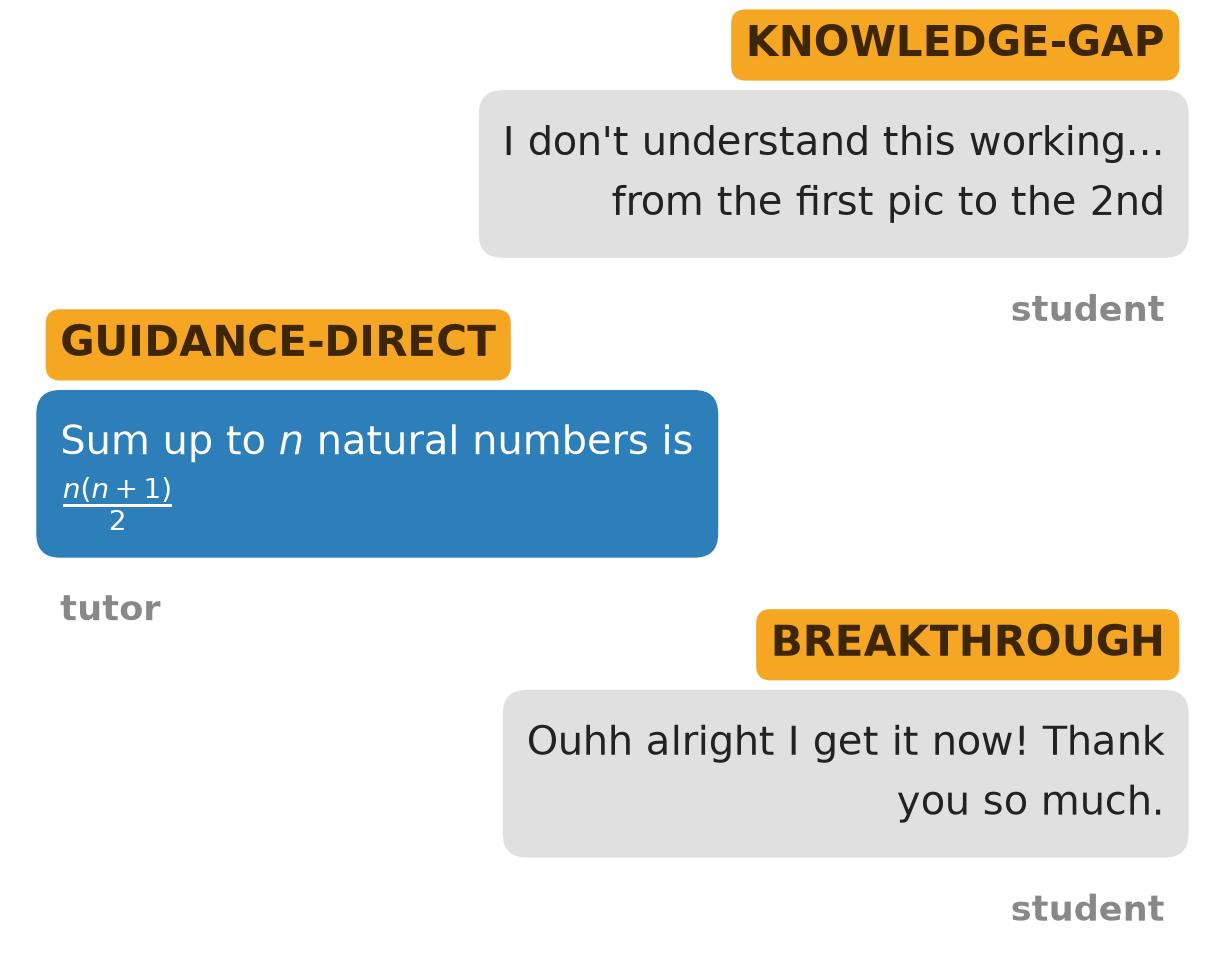}
\end{minipage}
\caption{\textbf{Left:} summary statistics for the two datasets. \textbf{Right:} a real excerpt from MathMentorDB's training split showing the tutor and student exchange, and the annotated move labels.}
\label{tab:dataset-summary}
\end{table}

\subsection{Metrics}
\label{sec:eval-metrics}

Our evaluation measures can be grouped along three axes: 1) \emph{Skeleton-level} metrics ($\mathrm{JS}_{\mathrm{marg}}$, $\mathrm{JS}_{\mathrm{trans}}$) test whether generated discourse-move trajectories reproduce real move dynamics; 2) \emph{Realized-conversation} metrics (pMSE, and the fidelity/coverage decomposition $\mathrm{IP}_\alpha$/$\mathrm{IR}_\beta$) test whether final transcripts match real conversations in a shared feature space; and 3) \emph{Authenticity} diagnostics test whether that match was novel or achieved by reproducing training records. We summarize the metrics below and refer the readers to Appendix~\ref{app:metrics} for specific details on their computation.

\paragraph{JS-Divergence on skeleton moves.} We report the Jensen--Shannon divergence over the move distributions $\mathrm{JS}_{\mathrm{trans}}$, averaging $\mathrm{JS}\bigl(\hat P_{\mathrm{gen}}(\cdot\mid c)\,\Vert\,\hat P_{\mathrm{real}}(\cdot\mid c)\bigr)$ over contexts $c=(z_{t-1},z_t,m_{t-1})$ to capture how well generated state sequences match real ones (lower is better).

\paragraph{pMSE and mode coverage on realized conversations.} To measure conversation \emph{realism}, we report mode \emph{coverage} and \emph{pMSE}, both computed on the per-conversation feature vector $\phi(x)$ of \S~\ref{sec:datasets}. For \emph{mode coverage}, we fit a $K$-mode Gaussian mixture on the held-out real split ($K{=}7$ tutoring, $K{=}8$ ESConv) and freeze it for evaluation. Each conversation is assigned a mode $\hat k(x)=\arg\max_k p(k\mid\phi(x))$ and use this to report \emph{coverage}, the number of modes with non-negligible mass among generated conversations. 

For \emph{pMSE}, we train an $L_1$-penalized logistic classifier to discriminate real from generated conversations and measure how far its out-of-fold propensity scores deviate from the no-signal baseline, following~\citet{ion2026measuring}. We report the normalized $\mathrm{pMSE}_{\mathrm{NORM}}\in[0,1]$, where $0$ means real and generated samples are indistinguishable (perfect fidelity) and larger values mean the classifier separates them (lower is better).

\paragraph{Sample-level fidelity, coverage, and authenticity.} pMSE reports whether a discriminator separates real from generated conversations, but not \emph{why}: the separation could be off-support generation, partial coverage of the real distribution, or memorized reuse of training data. We disentangle these with the three dimensions of \emph{fidelity, coverage and authenticity} \citep{alaa2022faithful}, which are all computed using on the same feature representation $\phi(x)$. \emph{Fidelity} ($\mathrm{IP}_\alpha$) is the fraction of generated conversations inside the ball containing the closest fraction $\alpha$ of real conversations; \emph{coverage} ($\mathrm{IR}_\beta$) measures the extent to which synthetic data spans the real data distribution; and \emph{authenticity} measures whether synthetic conversations stay as far from their nearest real training conversation as real conversations stay from each other. In the degenerate case, synthetic data can achieve high fidelity by copying training data points; fidelity and authenticity together ensure that generated points are both within distribution and novel. We report $\mathrm{IP}_\alpha,\mathrm{IR}_\beta\in[0,1]$ (higher is better; $1$ is a perfect distributional match), as well as the \emph{outlier} rate (synthetic conversations farther from real data than any real conversation).

\subsection{Baselines}
\label{sec:eval-baselines}

We compare against two baselines. 1) \textbf{RL+LLM} tests the effect of the GFlow
trajectory-balance (TB) loss: we train an architecture- and reward-matched policy
with REINFORCE~\citep{williams1992reinforce} and a mean-reward baseline in place of
TB. Comparing \gflowllm~with \rlllm~under
an identical reward isolates whether reward-proportional sampling from GFlowNets
prevents the diversity collapse motivated in \S\ref{sec:prelim}. 2) \textbf{LLM-only}
tests the effect of factored generation: a strong LLM generates complete
conversations end-to-end from the same framing and few-shot examples as
\S\ref{sec:realization}, across three tiers of increasing sophistication --
\textbf{B1} (naive): uniformly sampled few-shot examples; \textbf{B2}
(scenario-conditioned): additionally conditioned on a scenario bootstrap-resampled
from real conversation metadata; \textbf{B3} (mode-proportional): few-shot examples
selected in proportion to the same GMM mixture used by the \texttt{gmm\_density}
reward. We report the best-performing tier per domain in the main paper.




\subsection{Experimental details.} 
\label{sec:expdetails}
For each of the three synthetic conditions~\footnote{We identify the best performing LLM condition as \textbf{B2} in an aprior experiment with 200 samples.}, we generate $n{=}1500$ conversations, with \textsc{GPT-5.6-terra} and \textsc{Gemini-2.5-Flash} each; \gflowllm~and \rlllm~conditions realize a sampled skeleton (\S\ref{sec:realization}), while the LLM-only baseline produces structure and text jointly. Sample-level metrics compare equal-size sets, scoring 200 generated against 200 held-out real conversations per draw and reporting mean $\pm$ sd over repeated draws; because each draw uses 200 per side. All distances are Mahalanobis distances in $\phi(x)$, computed using normalized whitening of the feature values on the real training split. On the generated side, move labels are the skeleton the planner committed to (GFlowNet/RL) or the model's own tags (LLM-only), so move-rate features describe the \emph{planned} structure; whether the realized text follows that plan is not tested here. Full sampling, partition, and threshold details are in Appendix~\ref{app:eval-details}.


\input{5_eval}

\bibliography{iclr2027_conference}
\bibliographystyle{iclr2027_conference}

\appendix
\input{7_appendix}

\end{document}

%% file: math_commands.tex
\usepackage{amsmath,amsfonts,bm}

\def\eqref#1{equation~\ref{#1}}

\def\1{\bm{1}}

\DeclareMathAlphabet{\mathsfit}{\encodingdefault}{\sfdefault}{m}{sl}
\SetMathAlphabet{\mathsfit}{bold}{\encodingdefault}{\sfdefault}{bx}{n}



%% file: 3_method.tex
\section{Method}
\label{sec:method}
We generate synthetic conversations through a two-stage pipeline: (i)~a Generative Flow Network (GFlowNet) plans a sequence of speaker roles and abstract discourse moves underlying a conversation (discourse-move skeleton), and (ii)~a large language model realizes each move into a natural-language utterance. This separation lets the structural model learn from a compact, annotation-derived reward while leaving surface-form generation to a capable LLM.


We define the synthetic data generation framework (GFlowNet state, reward, and training objective over an abstract move set) (\S~\ref{sec:prelim}--\S~\ref{sec:method-abstract}), then instantiate and evaluate the framework on two domains of expert assistance - math tutoring, and emotional support dialogues (\S~\ref{sec:datasets}).

\subsection{Preliminaries: Generative Flow Networks}
\label{sec:prelim}

A Generative Flow Network (GFlowNet)~\citep{bengio2021flow} learns a stochastic policy for constructing a compositional object through a sequence of actions, trained so that the probability of sampling a complete object $x$ is proportional to a non-negative reward $R(x)$, rather than concentrated on the arg-max object as under reward-maximizing reinforcement learning. Construction is formalized as a walk on a directed acyclic graph whose nodes are partially built objects: a forward policy $P_F(\cdot \mid s;\theta)$ proposes the next action from state $s$, and generation terminates at a fully constructed object $x \in \mathcal{X}$. Among the family of objectives that enforce $P_F$-induced terminal probability proportional to $R(x)$ (flow matching, detailed balance, sub-trajectory balance) we use \emph{trajectory balance}~\citep{malkin2022trajectory}, which introduces a single learned scalar $Z_\theta$ approximating the partition function $\sum_x R(x)$ and matches, along each complete trajectory, the forward flow to the reward:
\begin{equation}
\mathcal{L}_{\text{TB}}(x;\theta) = \Bigl(\log Z_\theta + \sum_{t} \log P_F(a_t \mid s_{t-1};\theta) - \log P_B(a_t\mid s_t) - \log R(x)\Bigr)^2,
\label{eq:tb-general}
\end{equation}
with $P_B$ a (possibly fixed) backward policy. 

\subsection{Discourse-Move Trajectories as a Generative Flow Network}
\label{sec:method-abstract}

\paragraph{Trajectories.} We represent the conversation $\tau$ between an expert and a user as a sequence of speaker, move-label and utterance triplets ($z_t, m_t, u_t$) defined as $\tau~=~((z_0, m_0, u_0),(z_1, m_1, u_1),...,(z_t, m_t, u_t))$
where the speaker role $z_t \in \{\textsc{e},\textsc{c}\}$ is either an expert (E) or their counterpart (C). The moves $m_t$ belong to a domain-specific move vocabulary $\mathcal{M}_{\textsc{e}}$ and
$\mathcal{M}_{\textsc{c}}$. We write $\mathcal{M}=\mathcal{M}_{\textsc{e}}\cup\mathcal{M}_{\textsc{c}}$
for the full move set. The utterance $u_t$ is the observed natural language utterance of the speaker. 

\paragraph{Factored structural representation.}
We factor conversation generation into (i) generation of an interaction
structure and (ii) natural-language realization of that structure.
We define the structural action at turn $t$ as $a_t = (z_t, m_t)$ and represent the structural trajectory of a conversation as
\[
\tau_{STRUCT} = (a_1, a_2, \ldots, a_T)
     = ((z_1,m_1), (z_2,m_2), \ldots, (z_T,m_T)).
\]

The structural state at turn $t$ is the trajectory prefix $s_t = (a_1,\ldots,a_t)$, with
$s_0 = \varnothing$. Both our GFlowNet and reinforcement-learning conditions operate over
this structural state space rather than generating utterances directly. 

\paragraph{Policy.} For the policy, the speaker $z_t$ is
drawn from a fixed, data-derived speaker-transition distribution
$P_{\mathrm{role}}(z_t \mid s_{t-1})$ estimated from the corpus. Only the move policy is
learned, so the forward policy factors as
\begin{equation}
P_F(a_t \mid s_{t-1};\theta)
= P_{\mathrm{role}}(z_t \mid s_{t-1})\,\, P_F(m_t \mid s_{t-1}, z_t;\theta),
\label{eq:pf-factored}
\end{equation}
with the move factor $P_F(m_t \mid s_{t-1}, z_t;\theta)$ carrying learnable parameters
$\theta$. We use $P_F$ for this forward policy throughout, including in the
trajectory-balance objective and reward (Eq.~\ref{eq:reward-general}).

We parameterize
$P_F(m_t \mid s_{t-1}, z_t;\theta)$ with a GRU over the structural history: a recurrent state
$\mathbf{h}_t$ summarizes the prefix $s_t$, updated with the embedded (speaker, move) pair at
each step
(Appendix~\ref{app:policy}). Conditioning on the recurrent summary keeps the policy compact
while capturing longer-range structure than a one-step Markov policy.

\paragraph{Reward.} We factor the trajectory reward into a terminal, distributional
term and local, transition-level term:
\begin{equation}
\log R(\tau_{\mathrm{STRUCT}}) = \beta \log R_{\mathrm{term}}(\tau_{\mathrm{STRUCT}})
\;+\; \alpha \sum_{t=1}^{T} \log \bigl(r_t \cdot w_t\bigr).
\label{eq:reward-general}
\end{equation}
The terminal term $R_{\mathrm{term}}(\tau_{\mathrm{STRUCT}})$ scores the plausibility of
the \emph{whole} structural trajectory under a corpus-level feature map
$\phi:\mathcal{T}_{\mathrm{STRUCT}}\to\mathbb{R}^d$ (length statistics, per-move rates,
structural proportions, positional statistics, outcome indicators); by default we use a
Gaussian plausibility score
$R_{\mathrm{term}}(\tau_{\mathrm{STRUCT}})=\exp\!\bigl(-\lVert(\phi(\tau_{\mathrm{STRUCT}})-\mu)/\sigma\rVert_2\bigr)$
against the expert feature mean/std $(\mu,\sigma)$. The local term
$r_t=\max\bigl(\hat P_{\mathrm{data}}(m_t\mid m_{t-1},z_{t-1},z_t),\,\varepsilon\bigr)$ is
an empirical bigram transition probability estimated from the corpus (floored at
$\varepsilon$), rewarding locally coherent move sequences. 

\paragraph{Training.} We minimize the trajectory balance loss of Eq.~\ref{eq:tb-general} with the reward of Eq.~\ref{eq:reward-general}. Because a discourse-move skeleton is built by a single, monotonically growing sequence of moves, each state has a unique
parent, so the backward policy $P_B$ is degenerate and its term is a constant absorbed
into $\log Z_\theta$
We train in two regimes: an \emph{offline} regime that teacher-forces $P_F$ on real
training trajectories with $\log R(x)$ precomputed once per trajectory, and an
\emph{online} regime that samples fresh trajectories from the current policy, scores
them under $R$, and teacher-forces the same trajectories to obtain a differentiable
$\sum_t \log P_F$---closing the gap between the training distribution and the policy's
own sampling distribution. We train all models with Adam, gradient-norm clipping, and early stopping on a held-out split. \sumit{difficult to read, need to simplify}

\paragraph{Sampling and Natural-Language Realization}
\label{sec:sampling-abstract}
\label{sec:realization}

We generate conversations in two stages. \textbf{(1)~Skeleton sampling.} We first sample a move
skeleton: starting from a target length drawn from the corpus turn-count distribution, we
roll out $P_F$ turn by turn. At each step, we fix the next role from the speaker-transition
model, mask illegal moves, and sample the next move until a closing move or the target length is reached. 

\textbf{(2)~Realization.} We then pass the sampled skeleton sequence of (role, move) pairs to a large language model . The LLM model returns one
utterance per (role, move) pair as a structured object matching the fixed schema, resulting in the full natural language conversation.




%% file: 5_eval.tex
\section{Evaluation Results for GFlow Synthetic Data Generation}
\label{sec:results}
We now discuss results for our research questions on GFlowNet's suitability for diverse synthetic generation (RQ1), fidelity/coverage/authenticity against baselines (RQ2), and downstream task utility (RQ3). We refer to the three synthesis conditions as \textbf{\gflowllm, \rlllm, \llmonly}. 

\begin{table}[h]
\centering
\footnotesize
\setlength{\tabcolsep}{4pt}
\begin{tabular}{@{}l ccc ccc@{}}
\toprule
& \multicolumn{3}{c}{Math tutoring} & \multicolumn{3}{c}{ESConv} \\
\cmidrule(lr){2-4} \cmidrule(lr){5-7}
Condition & $\mathrm{JS}_{\mathrm{trans}}$ $\downarrow$ & \shortstack{pMSE $\downarrow$\\(all features)} & \shortstack{pMSE $\downarrow$\\(excl.\ surface)} & $\mathrm{JS}_{\mathrm{trans}}$ $\downarrow$ & \shortstack{pMSE $\downarrow$\\(all features)} & \shortstack{pMSE $\downarrow$\\(excl.\ surface)} \\
\midrule
Real conversations (null) & 0.063 & 2.0 & 1.7 & 0.003 & 4.6 & 3.2 \\
\cmidrule(lr){1-7}
\multicolumn{7}{@{}l}{\emph{GPT-5.6-terra realization}} \\
LLM only (\textbf{B2}, best tier) & 0.367 & 97.3 & 94.3 & 0.067 & 98.4 & 95.6 \\
RL + LLM (REINFORCE)$^\ddagger$ & 0.267 & 95.3 & 94.8 & 0.108 & 94.2 & 80.8 \\
GFlow + LLM (ours) & \textbf{0.145 }& \textbf{81.1} & \textbf{65.1} & \textbf{0.015} & \textbf{97.4} & \textbf{58.1} \\
\cmidrule(lr){1-7}
\multicolumn{7}{@{}l}{\emph{Gemini 2.5 Flash realization}} \\
LLM only (\textbf{B2}, best tier) & 0.363 & 99.3 & 92.8 & 0.130 & 94.0 & 82.7 \\
RL + LLM (REINFORCE)$^\ddagger$ & 0.267 & 97.1 & 96.5 & 0.108 & 90.2 & 81.7 \\
GFlow + LLM (ours) & \textbf{0.145} & \textbf{80.1} & \textbf{65.1} & \textbf{0.015} & \textbf{88.6} & \textbf{51.3} \\
\cmidrule(lr){1-7}
Structural upper bound$^\dagger$ & 0.065 & 44.5 & 2.7 & 0.004 & 72.6 & 6.1 \\
\bottomrule
\end{tabular}
\caption{GFlow achieved the best synthetic data fidelity across both datasets and LLM choices, as measured by state-transition fidelity ($\mathrm{JS}_{\mathrm{trans}}$) and pMSE across the three data generation conditions. A structural upper-bound is provided for reference (see \S~\ref{sec:expdetails} for sample sizes, redraws, and protocol). pMSE is reported on all features of $\phi(x)$ and with the surface feature group excluded, since surface features depend on the LLM realizer rather than the planned structure. Lower is better on all metrics.}
\label{tab:crossed}
\end{table}

\subsection{GFlow Suitability (RQ1) - GFlow Improves Synthetic Data Fidelity}
\label{sec:gflow-llm}

Table~\ref{tab:crossed} compares pMSE and $\mathrm{JS}_{\mathrm{trans}}$ across the three synthesis approaches: the best-performing end-to-end LLM baseline, RL+LLM, and GFlow+LLM. An optimal baseline computed using the real conversations is included for reference. Both factored approaches (RL and GFlow) optimize a Gaussian density reward derived from conversation features. Across both evaluation datasets, the GFlow+LLM pipeline outperforms all baselines on both metrics. 

\begin{figure}[h]
\centering
\begin{minipage}[b]{0.48\textwidth}
\centering
\includegraphics[width=\textwidth]{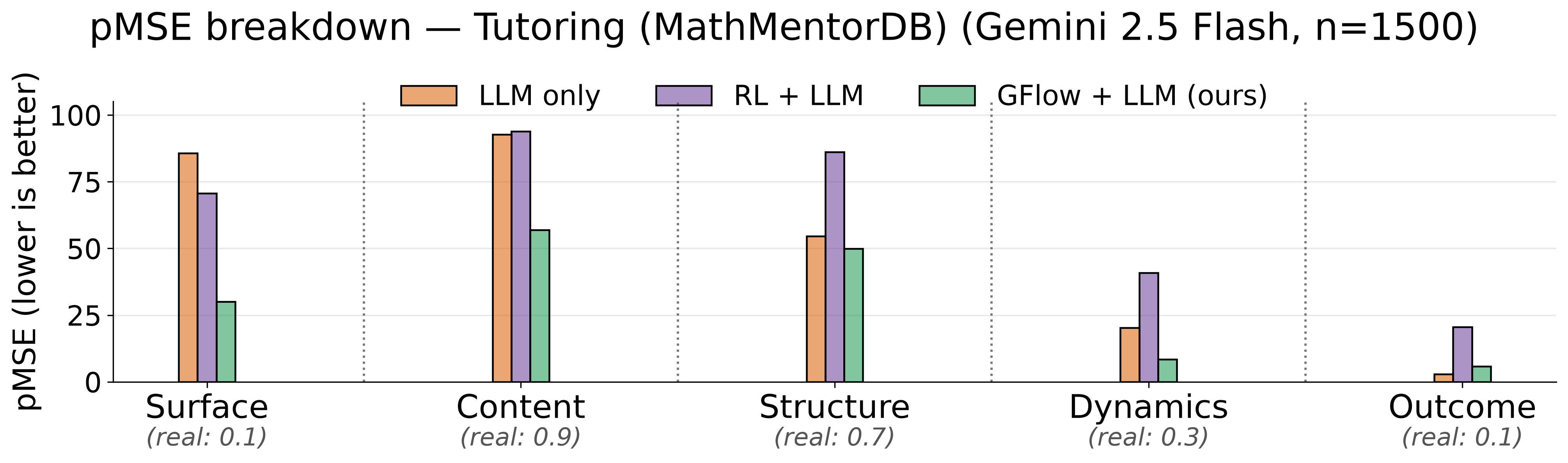}
\centerline{\small (a) Tutoring}
\end{minipage}
\hfill
\begin{minipage}[b]{0.48\textwidth}
\centering
\includegraphics[width=\textwidth]{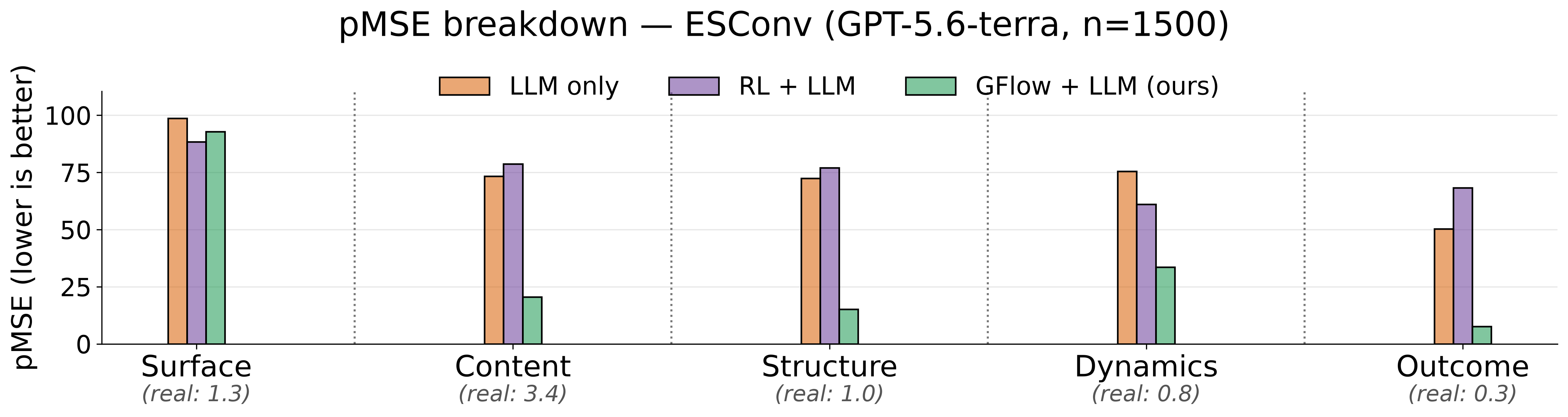}
\centerline{\small (b) ESConv}
\end{minipage}
\caption{A breakdown of pMSE results by feature dimension (\S\ref{sec:eval-metrics}) shows that GFlow+LLM superior results were achieved by consistent reduction of pMSE across all non-surface feature dimensions, which capture underlying conversation strategies.  Surface dimension pMSE was close across all three synthesis approaches. \gemini~
realization, $n{=}1500$, per domain. }
\label{fig:pmse-breakdown}
\end{figure}

Figure~\ref{fig:pmse-breakdown} breaks down these gains across five component dimensions (cognitive, content, structure, dynamics, and outcome; see \S~\ref{sec:datasets}). Factored GFlow yields substantial improvements across all five structural dimensions, while surface-level features achieve comparable pMSE across all conditions. This pattern aligns with the design of the generative models: surface features reflect the shared LLM generation step, whereas higher-level features depend on the latent conversation structure governed by the factored model. Details with individual LLM baseline tiers are provided in Appendix~\ref{app:llm-baseline}.


\subsection{Data Authenticity (RQ2) - GFlowNets balance fidelity and coverage while maintaining authenticity}
\label{sec:eval-copying}

\begin{table}[h]
\centering
\footnotesize
\begin{tabular}{@{}l ccccc@{}}
\toprule
& pMSE $\downarrow$ & $\mathrm{IP}_\alpha$ $\uparrow$ & $\mathrm{IR}_\beta$ $\uparrow$ & Auth.\ $\uparrow$ & Outlier \\
\midrule
Real held-out (reference) & 9.4{\tiny$\pm$4.5} & 0.94{\tiny$\pm$0.03} & 0.96{\tiny$\pm$0.02} & 0.70 & 0.050 \\
\midrule
LLM only (B2), GPT-5.6-terra & 95.0{\tiny$\pm$3.7} & 0.44{\tiny$\pm$0.04} & 0.09{\tiny$\pm$0.03} & 0.96 & 0.233 \\
REINFORCE, GPT-5.6-terra & 94.1{\tiny$\pm$4.1} & 0.81{\tiny$\pm$0.03} & 0.17{\tiny$\pm$0.04} & 0.92 & \textbf{0.006} \\
GFlowNet (TB), GPT-5.6-terra & \textbf{71.5{\tiny$\pm$3.3}} & \textbf{0.87{\tiny$\pm$0.04}} &\textbf{ 0.57{\tiny$\pm$0.07}} & 0.85 & 0.117 \\
\midrule
LLM only (B2), Gemini 2.5 Flash & 96.0{\tiny$\pm$2.6} & 0.21{\tiny$\pm$0.03} & 0.00{\tiny$\pm$0.01} & 0.99 & 0.738 \\
REINFORCE, Gemini 2.5 Flash & 93.4{\tiny$\pm$2.5} & 0.85{\tiny$\pm$0.02} & 0.16{\tiny$\pm$0.04} & 0.90 & \textbf{0.007} \\
GFlowNet (TB), Gemini 2.5 Flash & \textbf{70.6{\tiny$\pm$4.2}} & \textbf{0.89{\tiny$\pm$0.04}} & \textbf{0.59{\tiny$\pm$0.07}} & 0.84 & 0.117 \\
\bottomrule
\end{tabular}
\caption{Fidelity, diversity, and authenticity on MathMentorDB by synthesis condition. Metric definitions are in \S~\ref{sec:eval-metrics} and the sampling protocol is in \S~\ref{sec:expdetails}. The top row shows scores for held-out real conversations as if they had been generated, so it is the value a perfect generator reaches at this sample size. This provides an upper bound reference against which to compare the main results. ESConv results are in Appendix Table~\ref{app:tab:esconv-fidelity}.}
\label{tab:alpha-beta}
\end{table}

\begin{figure}[h]
\centering
\includegraphics[width=\textwidth]{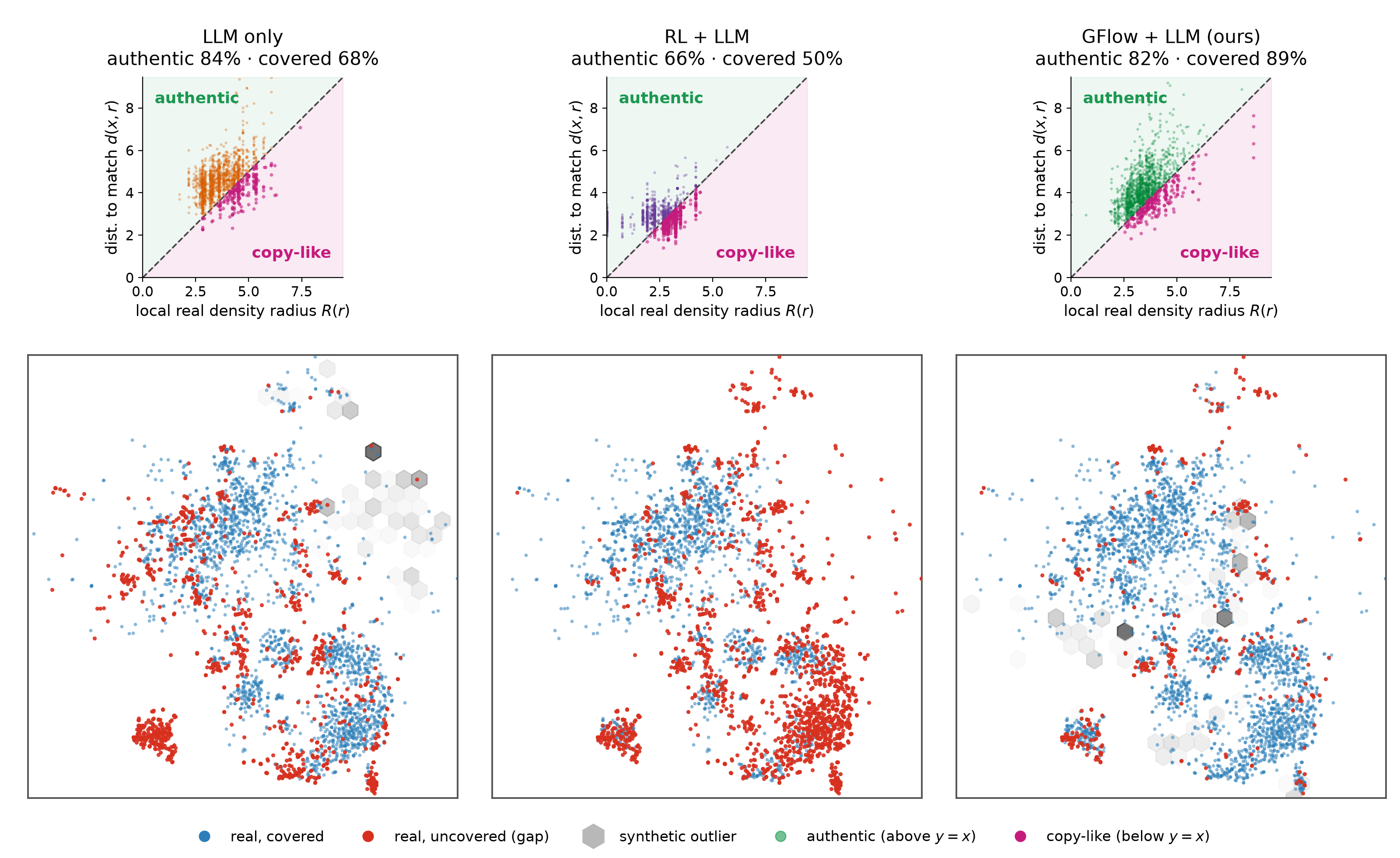}
\caption{Coverage of the real MathMentorDB distribution for the three studied synthesis approaches, showing how our GFlow + LLM approach (right) is the only one to achieve high fidelity coverage of the entire real support, while also producing novel but realistic outliers. Visualization uses t-SNE of non-surface $\phi(x)$ (based on Gemini, $n{=}1500$ per condition). A real point is covered if there is a synthetic point within the 95th percentile (\S~Appendix~\ref{app:eval-details}). Real conversations are blue if covered by the condition and red if not; grey hexagons are synthetic outliers. Plots above the panels describe the authenticity, a local density sensitive measure of how close are synthetic samples to real ones. A synthetic point is authentic if $d(x,r) > R(r)$ (above $y=x$ line), and copy-like otherwise. }
\label{fig:tsne-coverage}
\end{figure}

\paragraph{GFlow+LLM achieves the best balance of  coverage and authenticity.}
To understand whether \gflowllm~generates more realistic representations of real conversations, we report in Table~\ref{tab:alpha-beta} fidelity ($\mathrm{IP}_\alpha$), coverage ($\mathrm{IR}_\beta$), and copying statistics (authenticity, outlier) for every condition on MathMentorDB computed on the conversation features $\phi(x)$. For reference, we also report the values a perfect generator attains at this sample size. GFLow+LLM generation achieves competitive fidelity and coverage ($\mathrm{IP}_\alpha{=}0.87; \mathrm{IR}_\beta{=}0.57$) improving over the \rlllm~baseline by $0.06$ and $0.4$ respectively while maintaining moderate outliers and authenticity. 
\rlllm~achieves a high-fidelity but low coverage, by generating the dominant mode only. The \llmonly~baseline achieve low fidelity and coverage but high authenticity due to out-of-distribution examples that are far away from any real training data points.

To better illustrate these behaviors, Figure~\ref{fig:tsne-coverage} shows a visual representation of the synthetic data coverage on the real data for the three conditions, along with the separations between the synthetic and real data distributions.  LLM-only generations (left) are characterized by scattered coverage of the real data (67\%), with many outliers (45\%). RL+LLM generations (middle) are characterized by a skewed single mode coverage in the bottom right, but only covering half of the overall data manifold (50\%) and hardly any outliers (0.3\% outliers). GFLow+LLM generations are characterized by more uniform and better coverage of the data manifold (89\%), and spills out only modestly (14\% outliers) putting it in between LLM-only and RL+LLM conditions in terms of outliers. This illustrates the tradeoff between generating diverse synthetic data and avoiding outliers.

\subsection{Downstream utility (RQ3) - GFlow-generated diverse conversations provide richer downstream training signals}
\label{par:downstream-utility}
To show that diverse synthetic data can provide better signals for downstream tasks, we evaluate with a train-on-synthetic, test-on-real (TSTR) protocol~\citep{ramesh-etal-2025-synthtexteval} on the \mmdb~dataset. In this setup, we train a per-outcome logistic regression classifier on each turn's
sentence embedding fused with its discourse-move label,\footnote{Frozen
all-MiniLM-L6-v2~\citep{reimers-2019-sentence-bert} embeddings, mean-pooled over
three positional segments per conversation to retain coarse structure.} then
evaluate on held-out real conversations. We select three outcomes that characterize salient important aspects of effective tutoring interaction: 1) \emph{resolution} (did the student's confusion get resolved), 2) \emph{breakthrough} (did the student reach a moment of insight, a marker of genuine conceptual gain rather than mere answer delivery), and 3) \emph{off-topic drift} (did the conversation leave the instructional task, a marker of derailment). Since these outcomes can vary across modes, good performance on these tasks depends on having seen the full range of trajectories that experts can take.


\begin{table}[h]
\centering
\footnotesize
\resizebox{\linewidth}{!}{%
\begin{tabular}{@{}l c ccc ccc c@{}}
\toprule
& & \multicolumn{3}{c}{GPT-5.6-terra Realization} & \multicolumn{3}{c}{Gemini 2.5 Flash Realization} & \\
\cmidrule(lr){3-5}\cmidrule(lr){6-8}
Label & Real oracle & LLM & RL & GFlow & LLM & RL & GFlow & Majority floor \\
\midrule
\texttt{resolved} & 0.79{\tiny$\pm$0.03} & 0.50{\tiny$\pm$0.03} & 0.29{\tiny$\pm$0.01} & \textbf{0.65{\tiny$\pm$0.03}} & 0.57{\tiny$\pm$0.03} & 0.29{\tiny$\pm$0.01} & \textbf{0.64{\tiny$\pm$0.03}} & 0.37 \\
\texttt{has\_breakthrough} & 0.92{\tiny$\pm$0.02} & 0.71{\tiny$\pm$0.03} & 0.41{\tiny$\pm$0.01} & \textbf{0.72{\tiny$\pm$0.03}} & 0.69{\tiny$\pm$0.03} & 0.41{\tiny$\pm$0.01} & \textbf{0.81{\tiny$\pm$0.03}} & 0.41 \\
\texttt{has\_off\_topic} & 0.89{\tiny$\pm$0.04} & 0.48{\tiny$\pm$0.00} & 0.48{\tiny$\pm$0.00} & \textbf{0.50{\tiny$\pm$0.03}} & 0.48{\tiny$\pm$0.00} & 0.48{\tiny$\pm$0.00} & 0.48{\tiny$\pm$0.00} & 0.48 \\
\bottomrule
\end{tabular}%
}
\caption{$\mathrm{F1}_{\mathrm{macro}}$ results for three conversation outcome prediction tasks, using the train-on-synthetic test-on-real (TSTR) in the tutoring domain. GFlow + LLM wins or ties every (label, realizer) pair. $\pm$ is the half-width of a 95\% bootstrap CI (300 draws over real test conversations).}
\label{tab:tstr-text}
\end{table}

\begin{figure}[h]
\centering
\begin{minipage}[t]{0.49\textwidth}
\centering
\includegraphics[width=\textwidth]{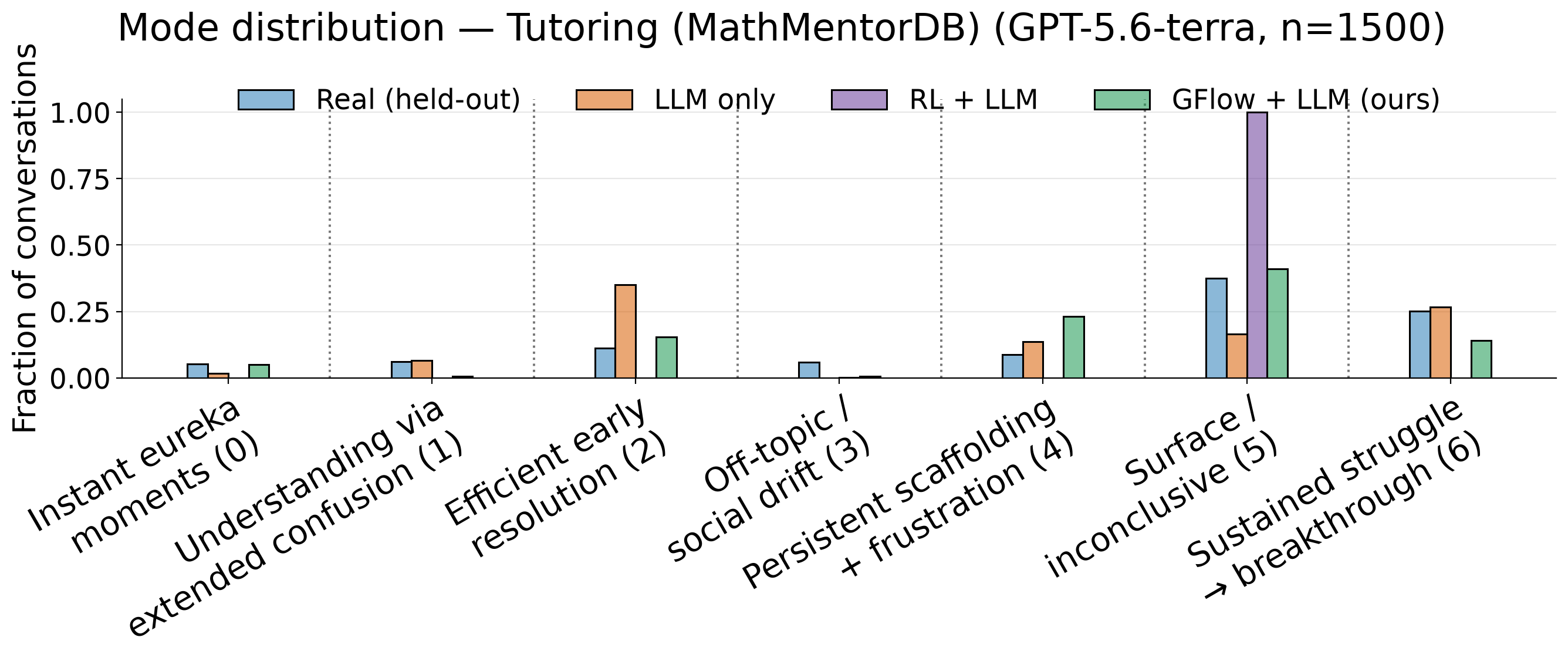}
\centerline{\small (a) Tutoring (7 modes)}
\end{minipage}
\hfill
\begin{minipage}[t]{0.49\textwidth}
\centering
\includegraphics[width=\textwidth]{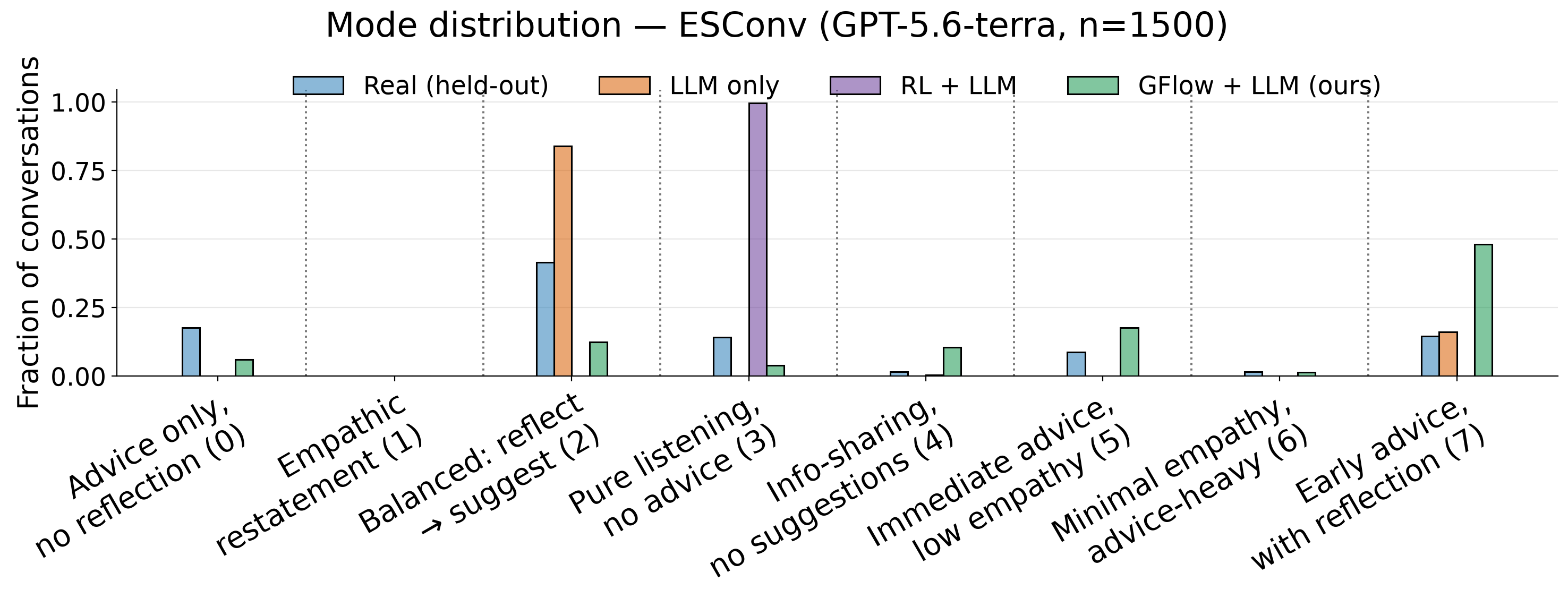}
\centerline{\small (b) ESConv (8 modes)}
\end{minipage}
\caption{GMM mode distribution on the real held-out set for the conditions \llmonly~(\textbf{B2}, best tier), \rlllm, and \gflowllm~(ours) (\S~\ref{sec:eval-metrics}). Mode labels are our post-hoc interpretation of each cluster's dominant feature profile. \rlllm~is characterized by a single dominant mode, and \gflowllm~reproduces more modes than \llmonly.}
\label{fig:mode-dist}
\end{figure}

\paragraph{Diverse conversations from GFlow+LLM improve training signals.}
Table~\ref{tab:tstr-text} reports the results for the outcome prediction task for the three synthesis approaches on both LLM realizations. We also include an oracle trained on the real conversations, and a majority class predictor as references. \gflowllm~beats the baselines for both LLM realizations, except for \texttt{has\_off\_topic} under Gemini realization. For this outcome, all three conditions degenerate to the majority prediction suggesting no learning signals for the task. For other conditions, LLM-only ranks second, and RL+LLM is much worse for all conditions due to its mode-seeking behavior. Figure~\ref{fig:mode-dist} shows the mode distribution across the conditions suggesting that the improved predictive power of data from \gflowllm~arises from its better mode coverage than baselines.


\section{Conclusion and Limitations}

Our central finding is that by using reward proportional sampling, and appropriate structure based rewards, we can generate diverse, structurally faithful synthetic conversations without relying on copying training data. Moreover, our method is able to navigate effectively between the extremes of verbatim resampling of real conversational structure, which has lowest detectability but high plagiarism rate, and end-to-end LLM generation that produces novel instances but fails to provide adequate coverage of real support. By adapting the GFlowNet framework to the problem of high-quality conversation synthesis we obtain synthetic data that is novel but that retains high fidelity and coverage of key modes.

Our work opens several interesting directions of research in high quality synthetic data generation. First, by controlling the reward structure that evaluates conversation, we can shape generation to a downstream task's specific needs. Rewards can also be informed from expert knowledge. For example, upweighting trajectories with prolonged confusion yields many conversation examples in which a student struggles before reaching understanding. Second, our generative approach is complementary to post-hoc discriminative methods that filter or rejection-sample synthetic data by quality~\citep{seedat2023curated}. These two approaches act on different synthesis stages and address different failures: generation determines which interactions exist in the corpus (coverage), while filtering determines which of them are kept (per-sample quality). A promising direction is to tune the two jointly: the discriminator's judgments can serve as a reward signal for the generator, while the generator learns to cover the space in ways the discriminator accepts rather than having its output pruned after the fact.

\paragraph{Limitations.} Our study covers two domains, both equipped with an established
discourse-move taxonomy. Extending the framework to other domains will require creating new discourse moves. The feature map $\phi$ is hand-designed, and required expert knowledge to create. Future work can investigate automatically learning features if more data is available. Our downstream evaluation trains classifiers on synthetic data as an initial step, but we leave full generative post-training for future work. We also acknowledge that a move-label abstraction is not a full representation of expertise. Exploring what drives these moves and how experts realize them are additional interesting future directions.

\paragraph{Reproducibility.} We provide code for reproducibility in the supplementary materials.




%% file: 7_appendix.tex
\section{LLM-use}
We used LLMs for generating synthetic data because that is the topic of the work. We used LLM-as-a-judge for analyzing the conversation modes and labeling the ESConv features similar to how features were labeled in MathMentorDB (MMDB) dataset. To achieve this, we used Claude to identify expert features in ESConv dataset that are analogous to MathMentorDB dataset such as \texttt{question\_rate}, \texttt{self\_disclosure\_rate}. After an initial taxomy generation, the features were validated by one of the authors for their correctness and relevance to the emotional support domain in accordance with liteature in the space. Some AI-assistance was part of implementation as we used AI-auto-complete enabled code editors for implementation. We also used LLMs to refine the arguments in the writing. We did not used LLMs to come up with ideas for the work. 

\section{Ethical Considerations}
\label{app:ethics}

We use two existing research datasets, MathMentorDB~\citep{ion2026measuring} and ESConv~\citep{liu-etal-2021-towards}, under their respective terms of use, and collect no new human data. One motivation for synthetic data is to reduce reliance on sensitive real conversations, but generation from real data does not by itself guarantee privacy. For this reason we report copying diagnostics (\S\ref{sec:eval-copying}), which show near-copy rates at or below those of held-out real data. These diagnostics do not constitute a formal privacy guarantee, and settings that require one should combine our approach with techniques such as differentially private training.

Both domains involve potentially vulnerable users, including students and people seeking emotional support. Synthetic conversations that are realistic by our metrics are not validated as safe or clinically appropriate. Systems trained on them should undergo domain-expert review and appropriate evaluation before deployment, particularly in mental-health settings, where synthetic data is not a substitute for professional oversight.

Because our generator reproduces the structural distribution of its training corpus, it also reproduces that corpus's biases, including which interaction patterns and user populations are represented. Reward reshaping can shift this distribution deliberately, which is useful for covering rare cases but could also be used to generate misleading or manipulative interactions. We will release code and data for research purposes and encourage users to document the reward and data choices behind any synthetic corpus they produce.

\section{Full Dataset Features}
\label{app:dataset-feature-set}

Table~\ref{app:dataset-feature-set} describes the full feature set used for both datasets.

\begin{table}[h]
\centering
\scriptsize
\begin{tabular}{@{}p{0.10\linewidth} p{0.43\linewidth} p{0.43\linewidth}@{}}
\toprule
\textbf{Dimension} & \textbf{MathMentorDB (tutoring), 47 feat.} & \textbf{ESConv (emotional support), 30 feat.} \\
\midrule
Surface (6 / 6) &
\texttt{avg\_student\_len, std\_student\_len, avg\_tutor\_len, std\_tutor\_len, n\_messages, turn\_ratio} &
\texttt{avg\_seeker\_len, std\_seeker\_len, avg\_supporter\_len, std\_supporter\_len, n\_messages, turn\_ratio} \\
\addlinespace
Content (21 / 8) &
\texttt{direct\_rate, scaffolding\_rate, correction\_rate, confirm\_pos\_rate, confirm\_neg\_rate, tutor\_question\_rate, knowledge\_check\_rate, attempt\_rate, knowledge\_gap\_rate, understanding\_rate, breakthrough\_rate, explain\_problem\_rate, explain\_reasoning\_rate, student\_question\_rate, student\_confirm\_rate, frustration\_rate, encouragement\_rate, empathy\_rapport\_rate, confidence\_express\_rate, acknowledgment\_rate, knowledge\_recall\_rate} &
\texttt{question\_rate, restatement\_rate, reflection\_rate, self\_disclosure\_rate, affirmation\_rate, suggestion\_rate, information\_rate, others\_rate} \\
\addlinespace
Structure (7 / 4) &
\texttt{pct\_tutor\_academic, pct\_student\_academic, pct\_non\_academic, pct\_socio\_emotional, greeting\_rate, platform\_rate, smalltalk\_rate} &
\texttt{pct\_empathic, pct\_directive, pct\_exploratory, pct\_personal} \\
\addlinespace
Dynamics (10 / 9) &
\texttt{n\_confusion\_episodes, avg\_confusion\_duration, max\_confusion\_duration, instant\_resolution\_rate, extended\_confusion\_rate, scaffolding\_density\_first\_third, scaffolding\_density\_last\_third, scaffolding\_density\_shift, first\_attempt\_position, first\_understanding\_position} &
\texttt{empathy\_density\_first\_third, empathy\_density\_last\_third, directive\_density\_first\_third, directive\_density\_last\_third, empathy\_to\_directive\_shift, strategy\_entropy, n\_strategy\_types, suggestion\_position, first\_empathy\_position} \\
\addlinespace
Outcome (3 / 3) &
\texttt{resolved, has\_breakthrough, has\_off\_topic} &
\texttt{resolved\_esc, has\_suggestion, has\_reflection} \\
\bottomrule
\end{tabular}
\caption{The full feature vector $\phi(x)\in\mathbb{R}^{47}$ (tutoring) / $\mathbb{R}^{30}$ (ESConv), by dimension (\S\ref{sec:method-abstract}), following the taxonomy of~\citet{ion2026measuring}. Per-dimension counts are (tutoring / ESConv). These are the only features used by any evaluation metric or reward in the paper (pMSE, mode coverage, DCR/copying, the GMM-density and Gaussian-$z$-score reward terms); the state trackers that condition the policy during generation (Appendix~\ref{app:policy}) are a separate, much smaller set and are not part of $\phi(x)$.}
\label{app:tab:feature-sets}
\end{table}

\section{GFlow and RL Policy Parameterization}
\label{app:policy}

We parameterize the move policy $P_F(m_t \mid s_{t-1}, z_t;\theta)$ of Eq.~\ref{eq:pf-factored}
with a GRU over the structural history. A recurrent state $\mathbf{h}_{t-1}$ summarizes the
(role, move) pairs seen through turn $t-1$, updated at each step with the embedded pair
$[\mathbf{e}(z_{t-1});\mathbf{e}(m_{t-1})]$. The role $z_t$ itself is not predicted by this
network -- it is drawn beforehand from the fixed $P_{\mathrm{role}}$ of Eq.~\ref{eq:pf-factored}.
Given $z_t$ and $\mathbf{h}_{t-1}$, the move is sampled from a masked softmax over $z_t$'s legal
move set,
\begin{equation}
P_F(m_t \mid s_{t-1}, z_t;\theta) = \operatorname{MaskedSoftmax}_{\mathcal{M}_{z_t}}\bigl(W_m[\mathbf{h}_{t-1};\mathbf{e}(z_t)] + b_m\bigr).
\end{equation}
When a Markov policy suffices, $\mathbf{h}_{t-1}$ is replaced by the embedded (role, move) pair
from turn $t-1$ alone and a projection of each active state tracker, concatenated through a
shallow MLP.



\section{LLM Baseline Condition Details}
\label{app:llm-baseline}

This appendix expands the baseline and reference conditions used in
Tables~\ref{tab:crossed}, and \ref{tab:reward-ablation}
(\S\ref{sec:eval-baselines}): the three LLM-baseline tiers, why their sample
size differs from the GFlowNet conditions, and a short glossary of the
RL+LLM, GFlow+LLM, and structural-upper-bound conditions.

\paragraph{LLM-baseline tiers.} The end-to-end LLM baseline of
\S\ref{sec:eval-baselines} is not one configuration but three escalating
tiers, each closing one specific objection to treating "prompt an LLM
directly" as a weak strawman.

\begin{itemize}
\item \textbf{B1 (naive).} Task specification plus $k{=}3$ few-shot
demonstrations drawn uniformly at random from the training corpus; no
scenario conditioning.
\item \textbf{B2 (scenario).} B0 plus conditional prompting: a
(attribute, value) scenario descriptor---resolution outcome and target turn
count for tutoring; emotional register and problem type and target turn
count for ESConv---bootstrap-resampled from an actual training
conversation's metadata, rather than an invented category list or a
uniform prior over labels.
\item \textbf{B3 (scenario+mode).} B1 plus mode-proportional few-shot
selection: the $k{=}3$ demonstrations are drawn with probability
proportional to the real GMM-mode mixture (\S\ref{sec:eval-metrics}),
reusing the identical mixture already fit for the \texttt{gmm\_density}
reward, instead of a uniform draw over training conversations. This hands
the LLM baseline the same distributional signal (real mode proportions)
that the GFlowNet's \texttt{gmm\_density} reward uses, so a residual
coverage gap in Table~\ref{tab:crossed}/Figure~\ref{fig:mode-dist} is
evidence of reward-proportional \emph{sampling}, not of withholding
information from the baseline.
\end{itemize}

In every tier the LLM self-tags: each output turn already carries a
\texttt{move} field the model chooses from the domain's legal per-role move
set, so structural fidelity is measured on a label the model committed to
itself, not one assigned by a separate classifier; the one role/domain
combination with only a single legal value (ESConv's
\texttt{Seeker-Response} sentinel) is asserted programmatically rather than
left to the model. Generation uses the same $k{=}3$ few-shot cardinality
and structured-output schema as our own realization pipeline
(\S\ref{sec:realization}), with a JSON-escape repair retry for occasional
malformed output.

\section{Evaluation Protocol Details}
\label{app:eval-details}

\paragraph{Real-side sampling and partitions.} Sample-level metrics compare 200 generated
conversations (drawn from the 1500 per condition) against 200 held-out real conversations,
with a reference row scoring 200 held-out real conversations as if generated. For tutoring,
the evaluation and reference samples are disjoint draws from the 1000 held-out conversations.
ESConv has only 260 held-out conversations, so its evaluation sample is drawn from those and
its reference sample from 200 training conversations withheld for this purpose, leaving 840
for reward calibration. We redraw the real side 10 times for pMSE and 50 times for
$\mathrm{IP}_\alpha$, $\mathrm{IR}_\beta$, and authenticity, redrawing the reference row as
often as every other row, and report means with standard deviations.

\paragraph{Distances and thresholds.} We estimate the whitening transform on training
conversations only, never on the conversations being scored. $\mathrm{IR}_\beta$ uses the
fifth-nearest real neighbour as its radius. A generated conversation is an outlier when its
distance to the nearest training conversation exceeds the 95th percentile of that distance
among held-out real conversations---$6.52$ for tutoring, $4.90$ for ESConv. Outlier rates are
computed over all 1500 conversations per condition; the other columns use the 200 drawn each
round. The structural upper bound is scored under the same protocol at $n{=}200$, realized by
\textsc{Gemini-2.5-Flash} alone.

\section{Metric Estimation Details}
\label{app:metrics}

\paragraph{pMSE.} The discriminator (\S\ref{sec:eval-metrics}) is fit with 5-fold
cross-validation; each record's propensity score $\hat p_i$ is its out-of-fold
prediction, avoiding the optimistic bias of scoring a record with a model fit on it.

\subsection{Sample-Level Fidelity, Coverage, and Authenticity: Definitions}
\label{app:sample-metrics}

All three metrics are computed on $\phi(x)$ after a whitening transform fit on the real
training split, so no single feature dominates the distances.

\paragraph{Mode computation} \sumit{TODO: }

\paragraph{$\alpha$-precision and $\beta$-recall.} \emph{$\alpha$-precision} $P_\alpha$ is the
fraction of generated conversations that fall inside the ball containing the closest fraction
$\alpha$ of real conversations; \emph{$\beta$-recall} $R_\beta$ is its mirror on the real side,
a real conversation counting as covered when its nearest generated neighbor lies within its
$k{=}5$ nearest-real-neighbor radius (Eq.~9 of~\citealp{alaa2022faithful}). A generator matching
the real distribution yields $P_\alpha{=}\alpha$ and $R_\beta{=}\beta$ for all $\alpha,\beta$;
we summarize each curve by its integrated deviation from that diagonal,
\begin{equation}
\mathrm{IP}_\alpha = 1-2\!\int_0^1\!|P_\alpha-\alpha|\,d\alpha, \qquad
\mathrm{IR}_\beta = 1-2\!\int_0^1\!|R_\beta-\beta|\,d\beta,
\label{eq:ip-ir}
\end{equation}
evaluated on a 30-step grid, so $1$ is a perfect match. By Theorem~1 of~\citet{alaa2022faithful}
both curves lie on the diagonal \emph{iff} the generated and real distributions coincide.

\paragraph{Authenticity and copying.} High $\mathrm{IP}_\alpha$ certifies that generated
conversations lie on the real support but not that they are \emph{novel} points on it: a
generator replaying training conversations would score just as well. \emph{Authenticity} tests
this via the distance from each generated conversation to its nearest real \emph{training}
record (distance to closest record, DCR): a generated conversation is inauthentic if its DCR
falls below that training record's own nearest-neighbor distance, i.e.\ it sits closer to a
training point than real conversations typically do. We also report a \emph{near-copy rate}
(fraction with DCR below $\tau_{\mathrm{mem}}$) and \emph{outlier rate} (fraction with DCR above
$\tau_{\mathrm{out}}$), with per-domain thresholds calibrated so a held-out real sample attains
outlier rate $0.05$ by construction and sets the reference for the other two: $\tau_{\mathrm{mem}}$
is the 5th percentile of leave-one-out DCR within the training split, and $\tau_{\mathrm{out}}$
the 95th percentile of held-out-real-to-training DCR. This follows novelty-alongside-quality
reporting in candidate-generation GFlowNets~\citep{bengio2021flow}. For each synthetic conversation, we plot its distance to its nearest real training match, $d(x,r)$, against that match's own nearest real neighbor $R(r)$~\citep{alaa2022faithful}.

\paragraph{Near-copy and outlier thresholds.} $\tau_{\mathrm{mem}}$ is the 5th percentile of leave-one-out DCR within the training split; $\tau_{\mathrm{out}}$ is the 95th percentile of held-out-real-to-training DCR. Both are calibrated per domain so that a held-out real sample, scored against the training split exactly as any synthetic condition is, has outlier rate 0.05 by  construction -- giving every other condition's near-copy and outlier rates a shared, interpretable reference point.

\subsection{pMSE for LLM-baseline conditions}
Table~\ref{app:tab:llm-tier-breakdown} summarizes the pMSE scores for the LLM conditions.

\begin{table}[]
\centering
\footnotesize
\begin{tabular}{@{}l cccccc@{}}
\toprule
Tier & Combined & Surface & Content & Structure & Dynamics & Outcome \\
\midrule
Tutoring B1 (naive) & 98.7 & 75.0 & 95.9 & 79.0 & 71.0 & 10.9 \\
Tutoring B2 (scenario) & 98.4 & 72.2 & 90.5 & 61.3 & 47.9 & 5.8 \\
Tutoring B3 (scenario+mode) & 98.6 & 70.3 & 89.5 & 68.3 & 44.8 & 6.8 \\
\cmidrule(lr){1-7}
ESConv B1 (naive) & 95.2 & 96.1 & 75.3 & 69.3 & 73.6 & 68.3 \\
ESConv B2 (scenario)$^\dagger$ & 95.0 & 87.2 & 62.4 & 68.1 & 70.4 & 51.8 \\
ESConv B3 (scenario+mode) & 96.8 & 98.1 & 78.8 & 69.7 & 72.5 & 53.4 \\
\bottomrule
\end{tabular}
\caption{Per-tier, per-feature-group pMSE for the three LLM-baseline tiers
(\S\ref{sec:eval-baselines}), $n{=}400$ each, GPT-5.6-terra.}
\label{app:tab:llm-tier-breakdown}
\end{table}

\subsection{Fidelity, Diversity, Coverage for ESConv}

\begin{table}[h]
\centering
\footnotesize
\begin{tabular}{@{}l cccccc@{}}
\toprule
Condition & pMSE $\downarrow$ & $\mathrm{IP}_\alpha$ $\uparrow$ & $\mathrm{IR}_\beta$ $\uparrow$ & Auth.\ $\uparrow$ & Near-copy & Outlier \\
\midrule
Real held-out (reference) & 7.0 {\scriptsize$\pm$1.6} & 0.96 {\scriptsize$\pm$0.01} & 0.96 {\scriptsize$\pm$0.01} & 0.67 & 0.072 & 0.050 \\
\cmidrule(lr){1-7}
LLM only (B2), GPT-5.6-terra & 95.5 {\scriptsize$\pm$2.3} & 0.43 {\scriptsize$\pm$0.03} & 0.06 {\scriptsize$\pm$0.02} & 0.94 & 0.000 & 0.123 \\
REINFORCE, GPT-5.6-terra & 96.0 {\scriptsize$\pm$1.6} & 0.27 {\scriptsize$\pm$0.02} & 0.06 {\scriptsize$\pm$0.02} & 0.96 & 0.001 & 0.282 \\
GFlowNet (TB), GPT-5.6-terra & 93.8 {\scriptsize$\pm$2.7} & 0.29 {\scriptsize$\pm$0.02} & 0.22 {\scriptsize$\pm$0.03} & 0.93 & 0.001 & 0.410 \\
\cmidrule(lr){1-7}

LLM only (B2), Gemini 2.5 Flash & 91.6 {\scriptsize$\pm$2.0} & 0.78 {\scriptsize$\pm$0.04} & 0.42 {\scriptsize$\pm$0.04} & 0.67 & 0.025 & 0.013 \\
REINFORCE, Gemini 2.5 Flash & 92.7 {\scriptsize$\pm$2.0} & 0.52 {\scriptsize$\pm$0.04} & 0.12 {\scriptsize$\pm$0.02} & 0.75 & 0.025 & 0.103 \\
GFlowNet (TB), Gemini 2.5 Flash & 86.4 {\scriptsize$\pm$3.4} & 0.52 {\scriptsize$\pm$0.04} & 0.50 {\scriptsize$\pm$0.05} & 0.86 & 0.005 & 0.249 \\
\bottomrule
\end{tabular}
\caption{Sample-level fidelity ($\mathrm{IP}_\alpha$), coverage ($\mathrm{IR}_\beta$), and authenticity for ESConv (\S\ref{app:sample-metrics}), $n{=}1500$ per condition, both realizers. Each draw scores 200 generated conversations against 200 held-out real conversations on $\phi(x)$, whitened on the training split; pMSE is the mean $\pm$ sd over 10 draws and $\mathrm{IP}_\alpha$, $\mathrm{IR}_\beta$, authenticity over 50 draws at $k{=}5$, with near-copy and outlier computed over all 1500. The first row scores held-out real conversations against the training split exactly as a synthetic condition is scored, giving the value a perfect generator attains at this sample size; read every other row against it rather than against $1$. pMSE here is $200$ vs.\ $200$ and so runs below the $n{=}1500$ values of Table~\ref{tab:crossed}, with the same ordering. The two providers used different ESConv realization prompts (\S\ref{sec:realization}), so rows compare within a provider; the GFlowNet rows are the \texttt{gmm\_density} reward with the inference-time directive mask off, not the guard-on configuration. \sumit{TODO: interpretation}}
\label{app:tab:esconv-fidelity}
\end{table}